\documentclass[11pt]{article}                  
\usepackage[paperheight=9.44in,paperwidth=6.69in,left=1.85cm,right=1.8cm,bottom=2.5cm,top=3cm,twoside]{geometry}

\usepackage[linesnumbered,ruled,vlined,boxed]{algorithm2e}
\usepackage{authblk}                              
\usepackage{fancyhdr}                             
\usepackage{lastpage}                             
\usepackage{textcomp}                             
\newcommand{\linia}{\noindent\rule{\linewidth}{0.5mm}\hrulefill} 
\usepackage{hyperref}

\SetAlFnt{\small}
\SetAlCapFnt{\small}
\SetAlCapNameFnt{\small}
\SetAlCapHSkip{0pt}
\IncMargin{-\parindent}

\usepackage{times}
\usepackage{titlesec}

\titleformat*{\section}{\large\bfseries}
\titleformat*{\subsection}{\normalsize\bfseries}

\fancypagestyle{firststyle}
{
\fancyhf{}
\setcounter{page}{53}
\fancyfoot{}
\fancyhead{}
\fancyhead[CE]{\upshape Zeszyty Naukowe WWSI, No 12, Vol. 9, 2015, pp. \thepage-\pageref{LastPage}}
\fancyhead[CO]{\upshape Zeszyty Naukowe WWSI, No 12, Vol. 9, 2015, pp. \thepage-\pageref{LastPage}}
\fancyfoot[L]{\footnotesize\bfseries Manuscript received April 11, 2015}
\fancyfoot[LE,RO]{\thepage}
}

\title{\large \bfseries Numerical Coding of Nominal Data} 
\author[1]{\normalsize Zenon Gniazdowski\thanks{E-mail: zgniazdowski@wwsi.edu.pl}}
\author[1]{\normalsize Micha{\l} Grabowski}
\affil[1]{\normalsize Warsaw School of Computer Science}

\date{\vspace{-5ex}}
\providecommand{\keywords}[1]{\textbf{\textit{Keywords ---}} #1}

\begin{document}

\maketitle                          
\thispagestyle{firststyle}          

\linia
\begin{abstract}
  \noindent In this paper, a novel approach for coding nominal data is proposed. For the given nominal data, a rank in a form of complex number is assigned. The proposed method does not lose any information about the attribute and brings other properties previously unknown. The approach based on these knew properties can been used for classification. The analyzed example shows that classification with the use of coded nominal data or both numerical as well as coded nominal data is more effective than the classification, which uses only numerical data.
\end{abstract}

\section{Introduction}
\label{Section:Introduction}
\noindent Different types of data are used in data analysis. Generally, they can be numerical data or nominal data. Numerical data are linearly ordered, which leads to the conclusion that two elements are equal or one element precedes the second one. Nominal data cannot be naturally ordered. In the set of nominal data, the identity equivalence relation can be defined, at most. It means that two elements may be equal or different.

For both types of data specific methods of analysis are developed. Particular difficulties arise when continuous and nominal data are analyzed simultaneously. Usually, by discretization continuous data are treated as nominal data. In this way, there avoids the opportunity of setting the data in order. On the other hand, the procedure can be reversed. In this case, nominal data are coded with the use of numbers \cite{Gra13}. Unfortunately, numerically coded nominal data cannot be naturally ordered.

In this paper, a novel approach for coding nominal data with the use of complex numbers will be presented \cite{Gni14}. For the given nominal data, it will be assigned a rank in a form of number. Proposed approach can be employed for classification and clustering.

\section{Preliminaries -- ranks of numerical data}
In statistics, in some cases, the numerical results of observations are replaced by their ranks. In the first stage of the ranking procedure, the data set is sorted in ascending order. Next, a rank equal to the item in the sorted set is assigned for the ordinal elements \cite{Wil45}. An example of this procedure is shown in Table \ref{Tab1}.

\begin{table}
\centering
\caption{Ranks of numerical data}\label{Tab1}
\begin{tabular}{c|c|c}
\hline
No. & Sorted data & Assigned rank\\
\hline\hline
1 & 21 & 1 \\
2 & 28 & 2 \\
3 & 33 & 3 \\
4 & 44 & 4 \\
5 & 45 & 5 \\
6 & 54 & 6 \\
7 & 55 & 7 \\
8 & 60 & 8 \\
9 & 63 & 9 \\
10 & 76 & 10 \\\hline
\end{tabular}
\end{table}

On the other hand, it may happen that in the sorted set there are different elements with the same values. In this case, the ranks assigned to identical values should be the same. Such elements receive rank that is equal to their average position in the sorted set \cite{Wil45}. These are so--called tied rank. Table \ref{Tab2} shows an example of tied ranks.

\section{Coding of nominal data}
As it is noted above, attaching ranks to numerical data is a function of the value of numerical data as well as the cardinality of its occurrences. In the case of nominal data, there is no differentiation of values. Possible method of ranking of nominal data has no chance to utilize the values. On the other hand, intuition suggests that in random community, both for numerical and nominal data, more numerous elements are more important than elements of less cardinality. Therefore, there is a proposal to rank of nominal data using only cardinality of identical elements.
 \subsection{Different frequencies of different nominal values}
 \label{}
 
 \begin{table}
 \centering
 \caption{Tied ranks of numerical data}\label{Tab2}
 \begin{tabular}{c|c|c}
 \hline
 No. & Sorted data & Assigned rank\\
 \hline\hline
 1 & 21 & 1 \\
 2 & 28 & 2 \\
 3 & \textbf{44} & \textbf{4} \\
 4 & \textbf{44} & \textbf{4} \\
 5 & \textbf{44} & \textbf{4} \\
 6 & 54 & 6 \\
 7 & \textbf{55} & \textbf{8.5} \\
 8 & \textbf{55} & \textbf{8.5} \\
 9 & \textbf{55} & \textbf{8.5} \\
 10 & \textbf{55} & \textbf{8.5} \\\hline
 \end{tabular}
 \end{table}

 Nominal data cannot be sorted according to their values, but it can be grouped according to identical values. In the $n$--element subset consisting of identical elements, these elements may be numbered from $1$ to $n$. For each of them can be assigned a rank that is equal to the average value of these numbers:
 \begin{equation}\label{Eq1}
 R=\frac{n+1}{2}
 \end{equation}
 
 More numerous elements will have higher rank than less numerous elements. As an example, a set consisting of $15$ elements with nominal values $\{a,a,a,a,a,a,b,b,b,b,b,c,c,c,c\}$  is considered. The set can be divided into three subsets. Each subset is a class of equivalence, which contains identical elements. To each nominal value, respective rank was assigned, according to the formula (\ref{Eq1}). The result of the ranking is shown in Table \ref{Tab3}.

\begin{table}
\centering
\caption{Rank of nominal data -- differences in the cardinality of elements}\label{Tab3}
\begin{tabular}{c|c|c|c|c|c|c|c|c}
\hline
\multicolumn{3}{c|}{First value} & \multicolumn{3}{c}{Second value} &  \multicolumn{3}{|c}{Third value} \\
\hline
Value&Position &Rank&Value&Position &Rank&Value&Position &Rank\\
\hline\hline
a&1&3.5&b&1&3&c&1&2.5\\
a&2&3.5&b&2&3&c&2&2.5\\
a&3&3.5&b&3&3&c&3&2.5\\
a&4&3.5&b&4&3&c&4&2.5\\
a&5&3.5&b&5&3&&&\\
a&6&3.5&&&&&&\\
 \hline
\end{tabular}
\end{table}

\subsection{The same frequencies of different nominal values}

For equinumerous subsets, method (\ref{Eq1}) gives the same rank. This would lead to the situation in which equinumerous elements will be indistinguishable. Therefore, method (\ref{Eq1}) should be modified. If the variable has several equinumerous subsets, the nominal elements belonging to the $j-th$ subset $(j = 0,1,\ldots,k-1)$ can be coded with the use of $k$ successive roots of unity:
 \begin{equation}\label{Eq2}
 R_j=R\cdot\sqrt[k]{-1}=R\cdot e^{i\phi}=R\cdot (\cos{\phi}+i\sin{\phi})
 \end{equation}

\begin{table}
\centering
\caption{Ranks of nominal data –- possible equinumerosity of elements in subsets}\label{Tab4}
\begin{tabular}{c|c|c|c|c|c}
\hline
 & Nominal & Module & Phase & Exponential  & Algebraic \\
No. & variable & $|R|$ & $ \phi [rad]$  & form $Re^{i\phi}$ &  form $a+bi$\\
\hline\hline
1 & a & 2 & 0 & 2 & 2\\
2 & a & 2 & 0 & 2 & 2\\
3 & a & 2 & 0 & 2 & 2\\
4 & b & 2 & $2\pi/3$ & $2e^{i2\pi/3}$ & $-1+1.73i$\\
5 & b & 2 & $2\pi/3$ & $2e^{i2\pi/3}$ & $-1+1.73i$\\
6 & b & 2 & $2\pi/3$ & $2e^{i2\pi/3}$ & $-1+1.73i$\\
7 & c & 2 & $4\pi/3$ & $2e^{i4\pi/3}$ & $-1-1.73i$\\
8 & c & 2 & $4\pi/3$ & $2e^{i4\pi/3}$ & $-1-1.73i$\\
9 & c & 2 & $4\pi/3$ & $2e^{i4\pi/3}$ & $-1-1.73i$\\
10 & d & 3.5 & 0 & 3.5 & 3.5\\
11 & d & 3.5 & 0 & 3.5 & 3.5\\
12 & d & 3.5 & 0 & 3.5 & 3.5\\
13 & d & 3.5 & 0 & 3.5 & 3.5\\
14 & d & 3.5 & 0 & 3.5 & 3.5\\
15 & d & 3.5 & 0 & 3.5 & 3.5\\
\hline
\end{tabular}
\end{table}

In the above expression $i=\sqrt{-1}$, $\phi=2\pi j/k$ $(j=0,1,\ldots,k-1)$ and $R$ is the rank calculated by the formula (\ref{Eq1}). Value of $\phi$ is the phase assigned to the successive $(j-th)$ nominal value. In the presented concept $R$ is a module of complex rank, depending on the cardinality of the subset that contains given nominal value. This approach gives the same modules $R$ for equinumerous subsets contained identical nominal elements, and distinguishes ranks of different nominal values via different phases.

Table \ref{Tab4} shows an example of the ranking for the case when the cardinality of elements $a$, $b$ and $c$ are equal to three, and the cardinality of the element $d$ is equal to six. For nominal values of $a$, $b$ and $c$ assigned phases are respectively equal to $0$, $2\pi/3$ and $4\pi/3$. Hence, the rank assigned to the value of $a$ is real, and ranks assigned to nominal values of $b$ and $c$ are complex. Real rank is assigned to the nominal value of $d$.

\section{Properties of complex coding}
In nominal data set, an equivalence relation can be defined for a data set in column corresponding to the given attribute. This relation divides this column into classes of equivalence. Each class will contain identical elements. The cardinality of each class is the only attribute information that is important from our analysis point of view. Coding with the use of complex numbers is unambiguous, i. e. after coding different elements are still distinguishable. In addition, it is also possible to define the corresponding equivalence relation, which divides the set into classes of equivalences, with cardinality of each class as before coding.

Coding does not lose any information about the attribute. The coded data receives additional properties that enrich them. Before coding, the cardinality of the given value was as the external feature. Now, through the module, the cardinality is an inherent property of the coded value of the attribute. The module presents information about the statistical strength of a given subset of elements. The phase contains the information about the number of equinumerous classes. Additionally, coding with the use of complex numbers brings other properties previously unknown. Above all, on complex numbers all arithmetic operations can be performed. Objects in data space can be viewed as vectors in a complex space. In this space, a scalar product, norm, as well as metric can be defined \cite{Kre67}. Scalar product of two complex vectors $x$ and $y$ is defined as follows:
\begin{equation}\label{Eq3}
 (x,y)=\sum_{i=1}^n x_i\overline{y}_i
 \end{equation}
 This way the norm, which is generated by the above scalar product, can also be defined:
\begin{equation}\label{Eq4}
||x||=\sqrt{(x,x)}
\end{equation}
By using this norm, metric also can be defined:
\begin{equation}\label{Eq5}
\rho (x,y)=||y-x||
\end{equation}

Proposed methodology of coding can be employed for analysis of nominal data. In particular, in a natural way it may be used for clustering and classification, because of the metric defined above.

\section{Possible application of proposed approach -- an example}
 The proposed approach can be used in $k$-means method, since the distance between objects in data space can be calculated by taking into account complex ranks of data. Data from the company that sells cars will be shown as an example \cite{Rut12}. The data set consists of ten objects (Table \ref{Tab5}). Each object is described by means of the seven attributes. The first two attributes (number of doors, engine power) are numbers. The other five attributes (color, fuel, interior, wheels and brand) take the nominal values. The first four of them were coded using the proposed approach (Table \ref{Tab6}).
  
 \begin{table}
 \centering
 \caption{Description of cars}\label{Tab5}
 \begin{tabular}{c|c|c|c|c|c|c|c}
 \hline
 No. & Door & Power & Color & Fuel & Interior & Wheel & Brand\\
 \hline\hline
1 & 2 & 60 & Blue & Petrol & Fabric & Steel & Opel\\
2 & 2 & 100 & Black & Diesel & Fabric & Steel & Nissan\\
3 & 2 & 200 & Black & Petrol & Leather & Alloy & Ferrari\\
4 & 2 & 200 & Red & Petrol & Leather & Alloy & Ferrari\\
5 & 2 & 200 & Red & Petrol & Fabric & Steel & Opel\\
6 & 3 & 100 & Red & Diesel & Leather & Steel & Opel\\
7 & 3 & 100 & Red & LPG & Fabric & Steel & Opel\\
8 & 3 & 200 & Black & Petrol & Leather & Alloy & Ferrari\\
9 & 4 & 100 & Blue & LPG & Fabric & Steel & Nissan\\
10 & 4 & 100 & Blue & Diesel & Fabric & Alloy & Nissan\\
\hline
\end{tabular}
\end{table}

The set of attributes will be divided into two subsets. The first six attributes are conditional attributes. The last one is an attribute of decision-making. Based on conditional attributes the set of objects will be clustered into three subsets. Afterwards, it is needed to check to what extent these subsets are consistent with the decision attribute. In other words, it is necessary to check whether the cars brand can be recognized from the car\textquotesingle s description. 
In order to verify the usefulness of the proposed complex coding, four classifications were made based on different conditional attributes:

 \begin{table}
 \centering
 \caption{Coded description of cars}\label{Tab6}
 \begin{tabular}{c|c|c|c|c|c|c|c}
 \hline
 No. & Door & Power & Color & Fuel & Interior & Wheel & Brand\\
 \hline\hline
1 & 2 & 60 & 2 & 3 & 3.5 & 3.5 & Opel\\
2 & 2 & 100 & -2 & 2 & 3.5 & 3.5 & Nissan\\
3 & 2 & 200 & -2 & 3 & 2.5 & 2.5 & Ferrari\\
4 & 2 & 200 & 2.5 & 3 & 2.5 & 2.5 & Ferrari\\
5 & 2 & 200 & 2.5 & 3 & 3.5 & 3.5 & Opel\\
6 & 3 & 100 & 2.5 & 2 & 2.5 & 3.5 & Opel\\
7 & 3 & 100 & 2.5 & 1.5 & 3.5 & 3.5 & Opel\\
8 & 3 & 200 & -2 & 3 & 2.5 & 2.5 & Ferrari\\
9 & 4 & 100 & 2 & 1.5 & 3.5 & 3.5 & Nissan\\
10 & 4 & 100 & 2 & 2 & 3.5 & 2.5 & Nissan\\
\hline
\end{tabular}
\end{table}

\begin{itemize}
\item Numbers and ad-hoc coded nominal data, 
\item Only numbers (number of doors and engine power), 
\item Only nominal data (color, fuel, interior, wheel) coded by the use of (\ref{Eq2}),
\item Numbers as well as nominal data coded by the use of formula (\ref{Eq2}).
\end{itemize}

$K$--means algorithm was used for classification. The data were standardized, for this purpose. Euclidean norm was used to measure distances. For this purpose, the adequate number of starting points for the $k$-means algorithm was chosen randomly. All these tests were repeated twenty times. In none of these twenty cases, the sequence of randomly selected points was not repeated.

After completion of the experiments, the results obtained for different conditional attributes were compared. Table \ref{Tab7} shows the comparison of classification results for different types of used data. It can be seen that the classification using coded nominal data and both numerical as well as coded nominal data is more effective than the classification, which uses ad hoc coding or only numerical data. Based on obtained results it must be concluded that the information that is contained in the coded nominal data is important for classification.

 \begin{table}
 \centering
 \caption{The results of classification}\label{Tab7}
 \begin{tabular}{c|c|c|c|c|c}
 \hline
 & \multicolumn{5}{|c}{Accuracy of classification}\\
 \cline{2-6}
Considered Data Type & $90\%$ & $80\%$ & $70\%$ & $60\%$ & $50\%$\\
\hline\hline
Ad hoc & -- & -- & -- & -- & 20\\
Only Numerical Data & -- & -- & 12 & 1 & 7\\
Only Coded Symbolic Data & 1 & 8 & 3 & 8 & --\\
Numerical and Coded Symbolic Data & 3 & 4 & 10 & 2 & 1\\
\hline
\end{tabular}
\end{table}

\section{Conclusions}
In this paper, a novel approach for coding nominal data was proposed. For the given nominal data, it can be assigned rank in a form of complex number. The module of this rank presents information about the statistical strength of a given subset of elements. The phase contains the information about the number of equinumerous values of attribute.

Proposed methodology is unambiguous. After coding, different values of attribute are still distinguishable. The method does not lose any information about the attribute. Additionally, coded data receives properties previously unknown that enrich them. Above all, on complex numbers all arithmetic operations can be performed. In complex space, a scalar product, norm, as well as metric can be defined. It means that coded data may be used for clustering and classification.

The well-known, folklore-type idea represents $m$--valued nominal domain as equidistant points in $R^m$ Euclidian space. Thus the classical Euclidean approach represents vectors of nominal values as points in $R^q$ space, where $q$ is equal to the sum of cardinalities of nominal domains under consideration. The proposed coding represents vectors of nominal values as points in $C^s$ space, where $s$ is equal to the number of nominal domains under consideration. We think that our coding can be considered as one of alternatives when the analyzed continuous--nominal data are sensitive to “course of dimensionality” (see \cite{Hastie2001}, \cite{Bel61}), due to low (compared to Euclidean coding) dimension of final space of codes. 

It is plain enough that the proposed coding injects additional information. For instance, the symmetries of codes of nominal data with equal frequencies are specific for the proposed coding schema and it may happen that the symmetries of the original data differ from the symmetries of their codes. Nevertheless, the other coding schemas involving nominal data frequencies information, for instance Bayesian coding of nominal values by scoring (see \cite{Kor08}), are subjected to this weakness as well. Moreover, Bayesian coding by scoring is applicable only to training data sets with two decision categories. Our proposal is fully general and can be applied to data sets with no decision categories information at all. 

The analyzed in this article data set shows, that classification with the use of coded nominal data or both numerical as well as coded nominal data is more effective than classification, which uses ad-hoc coding or only numerical data. From here, it must be concluded that the information that is contained in the coded nominal data is important for classification.

The presented proposal is preliminary proposal. Although the method looks interesting, further investigations of this approach are necessary, for instance a serious experimental study. These investigations could confirm the usefulness of the method. They could also show other possible applications of the proposed method.

\bibliography{NumericalCoding}
\bibliographystyle{unsrt}

\end{document}